# On the Application of Efficient Neural Mapping to Real-Time Indoor Localisation for Unmanned Ground Vehicles

Christopher J. Holder*
School of Computing
Newcastle University
Newcastle Upon Tyne, United Kingdom
chris.holder@newcastle.ac.uk
*Corresponding author

Muhammad Shafique
Division of Engineering
New York University Abu Dhabi
Abu Dhabi, United Arab Emirates
ms12713@nyu.edu

*Abstract—* **Global localisation from visual data is a challenging problem applicable to many robotics domains. Prior works have shown that neural networks can be trained to map images of an environment to absolute camera pose within that environment, learning an implicit neural mapping in the process. In this work we evaluate the applicability of such an approach to real-world robotics scenarios, demonstrating that by constraining the problem to 2-dimensions and significantly increasing the quantity of training data, a compact model capable of real-time inference on embedded platforms can be used to achieve localisation accuracy of several centimetres. We deploy our trained model onboard a UGV platform, demonstrating its effectiveness in a waypoint navigation task, wherein it is able to localise with a mean accuracy of 9cm at a rate of 6fps running on the UGV's onboard CPU, 35fps on an embedded GPU, or 220fps on a desktop GPU. Along with this work we will release a novel localisation dataset comprising simulated and real environments, each with training samples numbering in the tens of thousands.**

Keywords—navigation, neural networks, image processing

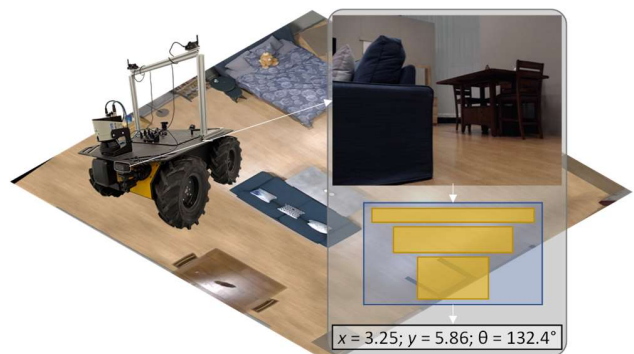

Figure 1. Overview of our deployed neural mapping system: A UGV mounted camera captures an RGB image, which is input to a trained neural network that predicts the absolute 2D pose of the UGV within the scene.

## I. INTRODUCTION

Global localisation, that is the retrieval of the absolute pose of an agent within a defined space, in GPS-denied environments is a challenging problem affecting a wide-array of robotics applications. Solutions that involve the placement of fixed markers or beacons, such as ultra-wideband positioning [1], ultrasonic tracking beacons [2] or visual markers [3] can facilitate accuracy ranging from centimetres to metres, and require specialist hardware be placed within the environment and in some cases on agents themselves. Vision-based approaches, wherein image and/or depth data captured by an agent's onboard sensors is compared to a pre-built map of an environment, have demonstrated significant potential as an alternative, and can broadly be divided into five categories as described in [4]:

1. In 3D – 3D localisation, pose is estimated via correspondence between 3D points captured by depth camera or LIDAR and those contained in a pre-built 3D map [5] [6] [7];

2. 2D – 3D Descriptor Matching approaches compute features for keypoints in captured images, which are subsequently matched to pre-computed descriptors contained within a 3D map [8] [9] [10];

3. 2D - 3D Scene Coordinate Regression estimates 3D coordinates for image pixels before performing 3D – 3D localisation [11] [12] [13];

4. 2D – 2D Explicit Map Localisation involves matching captured images to those in a database of images with known poses, then computing the relative pose against retrieved images to refine the final estimate [14] [15] [16];

5. 2D – 2D Implicit Map Localisation, that we refer to in this work as neural mapping, estimates pose via a neural network that has learned an implicit representation of a scene from a training set of image–pose pairs [17] [18] [19].

Neural mapping was first proposed in [17], in which a dataset of five outdoor scenes each comprising between 200 and 3000 training images was used to train a Googlenet [20] model to predict 7-dimensional poses in 3D space, resulting in a median position error between 1.4m and 3m and orientation error between 2.4° and 4°. The approach was also evaluated on the 7 Scene dataset [21], comprising seven indoor scenes each with between 1000 and 7000 RGB-D training images, demonstrating a median position error of between 0.3m and 0.6m and orientation error of between 3.3° and 7°. Numerous works have proposed improvements to the technique, including the augmentation of training data with synthetic images [22], exploiting temporal constraints in image sequences [23], geometry aware loss functions [24], joint prediction of pose and visual odometry between image pairs [25], masking of dynamic objects that could otherwise confound localisation [26], and joint prediction of pose and semantic segmentation [27]. *However, many of these techniques rely on large, complex models that are not feasible for deployment on mobile robotic platforms requiring real-time inference – e.g.* [27] *runs at 12fps on a desktop GPU, while our approach achieves 220fps when run on comparable hardware.*

In this work we aim to evaluate the viability of a simple, efficient neural mapping approach applied to a real-world robotics scenario. We demonstrate that by constraining the problem to 2D position and one axis of rotation, as would typically be the case for an unmanned ground vehicle (UGV) operating indoors, and increasing the magnitude of training data over previous datasets, highly accurate localisation is possible, even when a relatively compact model is used. We first validate our approach using three simulated environments in the Unity game engine [28], creating training sets of up to 200,000 samples to assess the impact of training data quantity, and train models of several different architectures and sizes and with different combinations of hyperparameters, assessing accuracy, model size and inference speed. We demonstrate a capability to achieve an average position error of <4cm with a model capable of running at 220fps on a desktop GPU.

We then capture data in a real environment using a UGV equipped with an RGB camera, creating a training set of 32,638 samples, and train models based on the same set of architectures to evaluate for discrepancy between the use of simulated and real datasets, and demonstrate the capabilities of the approach in real-world scenarios. Finally, as depicted in Figure 1., we deploy a trained model onboard a UGV and demonstrate the real-world applicability of our neural mapping approach in a simple waypoint navigation task, being, to our knowledge, the first work to do so. The novel dataset created as part of this work, comprising image - pose pairs from three simulated and one real environment will be made available for other researchers to evaluate their approaches.

**Proposed Contributions:**

1. A new dataset comprising image pose pairs captured in simulated and real environments;
2. A thorough evaluation of several CNN architectures in multiple configurations encompassing pose regression performance and runtime on various hardware platforms;
3. We demonstrate a capability to deploy a trained model on a UGV platform for real-time localisation within a realistic indoor navigation scenario.

## II. Neural Mapping

In this section, we describe the approach used to train our neural mapping models. The goal in training such a model is that it learn an implicit representation of an environment from a large number of image - pose pairs captured within that environment, allowing it to accurately estimate the corresponding pose of unseen images captured in the same environment, as illustrated in Figure 1.

Each model comprises a convolutional neural network that takes a single colour image as input, and estimates the absolute 2D coordinates and orientation, each normalised to -1 to +1, within the environment from which the image was captured. We evaluate five configurations of ResNet [29] architecture – ResNet-18, -34, -50, -101, and -152 – alongside VGG19 [30] and MobileNet-V2 [31], each first pretrained for an image classification task [32], and with the output classifier replaced by 3 regression outputs and a hyperbolic tangent regularisation function. Our models are trained using samples numbering in the tens or hundreds of thousands, to provide as comprehensive a coverage of the position – orientation space as possible, with each sample comprising a colour image paired with ground truth coordinates and orientation. L1 loss is computed over the three output variables – L1 was empirically found to facilitate greater accuracy than alternatives such as L2, possibly due to its capability to penalise small magnitude errors – and the Adam [33] optimiser is used to subsequently update model parameters. Learning rate decay is used to enable finer-grained optimisation as the model learns, however we found that by periodically resetting the learning rate the model demonstrated better overall results, likely due to an increased aversion to local minima. We use an exponential decay function with a rate of 0.9998 applied every 1000 iterations, with the learning rate reset to its initial value of $10^{-4}$ if no increase in accuracy is observed over 10,000 iterations. If no improvement is observed after a further 10,000 iterations we take the model to have converged and conclude training. We apply a weight decay of $10^{-6}$ and use a batch size of 32, evaluating models with input dimensions of 128×128, 256×256, and 512×512.

TABLE I. Details of the Environments in our dataset

| *Name* | *Type* | *Size* | *Samples* | *Example Image* |
|---|---|---|---|---|
| Cabin | Sim | 105m² | 200,000 | 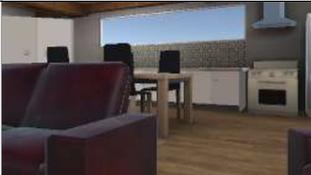 |
| Warehouse | Sim | 1200m² | 200,000 | 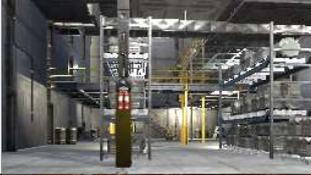 |
| Temple | Sim | 4200m² | 200,000 | 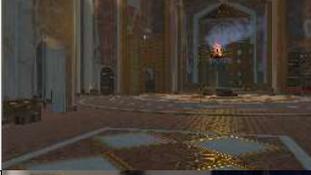 |
| Apartment | Real | 64m² | 32,638 | 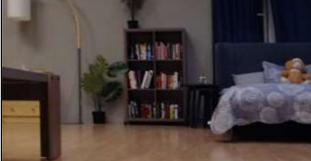 |

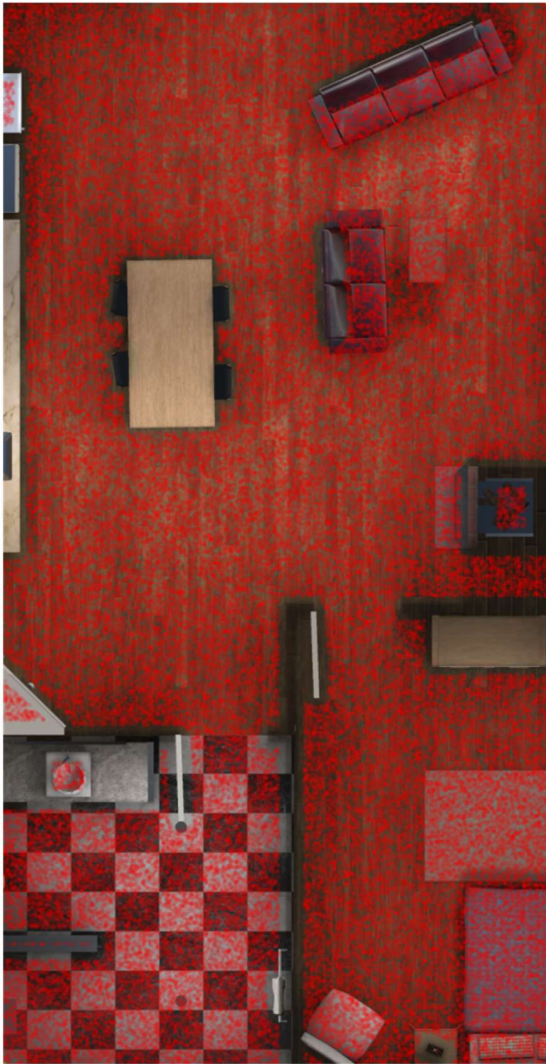

Figure 2. Map of our simulated Cabin environment, with each sample from the respective training set plotted in red.

## III. DATASET

### A. Simulated Environments

To validate our approach and identify the most effective training hyperparameters, we create three datasets in photorealistic simulated environments using the Unity game engine [28], each comprising 200,000 training samples and 500 test samples, ranging from $105m^2$ to $4200m^2$. For each sample, a random set of $x$ and $y$ coordinates and yaw angle are selected, and after a collision detection step to ensure the chosen location is not inside an obstacle, a camera is placed within the environment at a fixed height with zero pitch and roll to capture an image which is saved along with the $x$, $y$ and yaw values.

Table 1 details the environments used for our datasets. The smallest simulated environment represents the interior of a residential log cabin with dimensions of 7m × 15m, comprising a large living room and kitchen area with separate bedroom and bathroom. The Warehouse environment is set inside a science-fiction themed warehouse comprising storage and maintenance areas of 30m × 40m. The Temple environment covers 60m × 70m and includes interior and exterior areas surrounding a large medieval fantasy-style temple building.

Figure 2. shows our 100,000 sample training set plotted on a map of the Cabin environment, demonstrating comprehensive coverage of all free space within the scene.

### B. Real Environment

To demonstrate that our approach can be applied to real applications outside of simulation, we capture a dataset in a real-world environment using a UGV, a process that poses significant challenges compared to the simulated data, both in terms of the data collection process itself and in training an accurate model based on the resultant data. A key goal of the data collection process is to cover as much of the potential location/orientation space as possible, easily achieved in simulation but time consuming when manoeuvring a UGV around a real environment. Inaccurate or noisy ground-truth position and orientation data has the potential to severely impact the accuracy of the resulting trained model. While visual consistency can be an enforced constraint in simulated environments, real images can be impacted by lighting changes, rolling shutter effect as the camera moves, and marks left on the environment by the UGV itself, all of which have the potential to impact the representation of the environment learned by the model.

The environment used is a soundstage in a film studio arranged to resemble the interior of a small studio apartment with dimensions of 8m × 8m. We use a Clearpath Husky UGV [34], pictured in Figure 3., equipped with a webcam for image capture and a Marvelmind ultrasonic positioning system [2], which combined with IMU data is capable of providing ground-truth location and orientation data to within 2cm. This could potentially be improved with the use of visual odometry [35] or structure from motion [36] algorithms, however we were able to achieve satisfactory results using only the raw position data.

Samples are captured via a combination of autonomous and manual traversal of the environment. During autonomous navigation the UGV performs a random walk traversal of the environment, utilising a precomputed map of obstacles to avoid collisions, with a new sample captured each time the platform has moved > 10cm or rotated >10° since the last capture. We find this approach to provide good coverage of the environment, however due to a conservative approach to collision avoidance, we also capture data from a manual traversal of the environment wherein a human operator is able to manoeuvre the platform closer to obstacles to improve coverage of areas too risky for autonomous navigation. In total, this provides a dataset of ~33,000 samples, captured over approximately 10 hours of UGV navigation of our environment. Figure 4. Visualises the captured samples on a map of the environment, demonstrating the difference in coverage compared to the simulated environment shown in Figure 2.

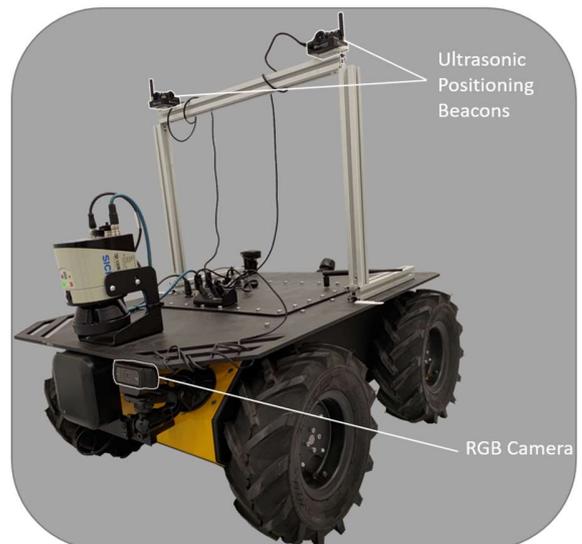

Figure 3. Husky UGV platform used for dataset collection, with sensors highlighted.

## IV. EVALUATION

### A. Simulated Environments

We evaluate our trained models over a test set of 500 samples from each simulated environment, quantifying the effects of model architecture, input dimensions, and number of training samples used.

TABLE II. ACCURACY OF DIFFERENT MODEL ARCHITECTURES ACROSS SIMULATED ENVIRONMENT TEST SETS

| Cabin | Position Error (m) | | | | Orientation Error (°) | | | |
|---|---|---|---|---|---|---|---|---|
| Model | μ | σ | Max | RMSE | μ | σ | Max | RMSE |
| *Resnet-18* | 0.053 | 0.032 | 0.291 | 0.062 | 1.012 | 2.638 | 45.60 | 2.826 |
| *Resnet-34* | 0.034 | 0.021 | 0.153 | 0.040 | 0.551 | 0.511 | 5.651 | 0.752 |
| *Resnet-50* | 0.037 | 0.025 | 0.162 | 0.044 | 0.524 | 0.499 | 4.564 | 0.724 |
| *Resnet-101* | 0.034 | 0.021 | 0.137 | 0.040 | 0.525 | 0.520 | 5.311 | 0.739 |
| *Resnet-152* | **0.030** | **0.020** | **0.123** | **0.036** | **0.439** | **0.427** | **3.240** | **0.612** |
| *VGG19* | 0.046 | 0.029 | 0.179 | 0.054 | 1.028 | 3.774 | 65.42 | 3.912 |
| *MobileNet-v2* | 0.071 | 0.044 | 0.290 | 0.083 | 1.169 | 1.132 | 12.39 | 1.628 |

| Warehouse | Position Error (m) | | | | Orientation Error (°) | | | |
|---|---|---|---|---|---|---|---|---|
| Model | μ | σ | Max | RMSE | μ | σ | Max | RMSE |
| *Resnet-18* | 0.369 | 1.071 | 16.11 | 1.132 | 1.879 | 6.257 | 89.88 | 6.534 |
| *Resnet-34* | **0.237** | 0.650 | 9.901 | 0.692 | **1.192** | 3.436 | 46.86 | 3.637 |
| *Resnet-50* | 0.354 | 0.840 | 11.45 | 0.912 | 1.940 | 6.019 | 92.14 | 6.324 |
| *Resnet-101* | 0.346 | 1.085 | 15.86 | 1.139 | 2.189 | 11.60 | 174.9 | 11.81 |
| *Resnet-152* | 0.287 | **0.518** | 7.191 | **0.592** | 1.233 | **2.303** | **34.12** | **2.613** |
| *VGG19* | 0.373 | 0.825 | 13.09 | 0.905 | 2.283 | 7.579 | 139.7 | 7.915 |
| *MobileNet-v2* | 0.509 | 0.632 | **5.894** | 0.812 | 3.601 | 11.21 | 157.1 | 11.77 |

| Temple | Position Error (m) | | | | Orientation Error (°) | | | |
|---|---|---|---|---|---|---|---|---|
| Model | μ | σ | Max | RMSE | μ | σ | Max | RMSE |
| *Resnet-18* | 0.701 | 2.364 | 35.74 | 2.465 | 2.325 | 8.907 | 132.3 | 9.205 |
| *Resnet-34* | 0.624 | 2.413 | 37.08 | 2.493 | **1.432** | 4.240 | 73.38 | **4.475** |
| *Resnet-50* | **0.533** | 2.023 | 40.51 | 2.092 | 2.025 | 7.363 | 93.22 | 7.636 |
| *Resnet-101* | 0.561 | 2.033 | 35.99 | 2.109 | 1.566 | 4.533 | 67.32 | 4.796 |
| *Resnet-152* | 0.643 | 2.596 | 38.30 | 2.674 | 1.616 | **4.205** | **63.01** | 4.504 |
| *VGG19* | 0.667 | **1.937** | **28.29** | **2.049** | 2.258 | 7.871 | 120.6 | 8.188 |
| *MobileNet-v2* | 0.718 | 2.058 | 36.64 | 2.179 | 2.488 | 7.474 | 93.27 | 7.877 |

#### 1) Network Architecture

We detail the performance of each model tested on the three simulated environments in Table 2, listing the mean, standard deviation, maximum and root mean squared error for predicted position, in metres, and orientation, in degrees.
In each case we use input dimensions of 256×256, a training set of 100,000 samples and a batch size of 32.

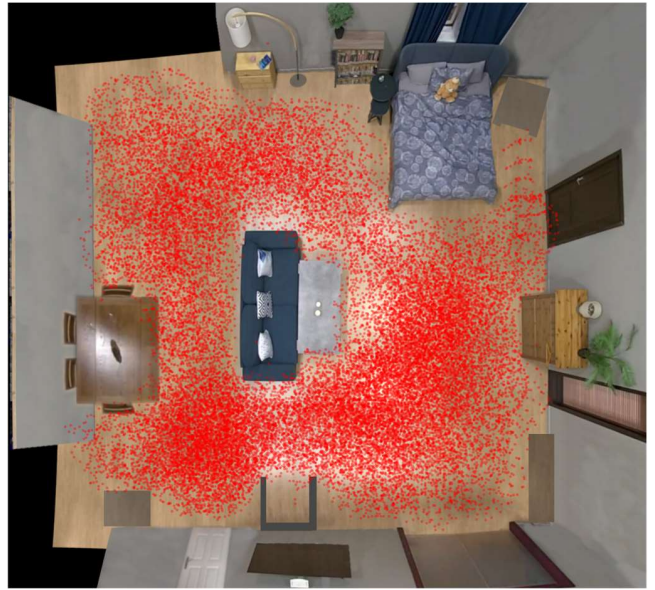

Figure 4. Top-down view of the real Apartment environment with all samples from its training data plotted in red. While coverage is good in open areas, compared to the simulated environment shown in Figure 2., there are fewer samples close to obstacles due to the difficulty of navigating in such areas.

It can be observed that in general, larger Resnet models perform better than their smaller counterparts, although this is not consistent throughout, while Mobilenet, a much more compact model, performs poorly. Despite comprising a much larger model, VGG19 does not demonstrate meaningful performance gains over Resnet, demonstrating the efficacy of residual connections. Figure 5. Visualises the samples of the Cabin test set and corresponding predictions output by the Resnet-101 model.

#### 2) Input Dimensions

Table 3 shows the results of our evaluation of the impact of input size on model accuracy, listing mean position error $P$ in metres and mean orientation error $\theta$ in degrees. In each case, a Resnet-101 model is trained and tested with input images of the listed dimensions, using a training set of 100,000 samples and a training batch size of 32. While a higher input resolution might be expected to facilitate the extraction of more fine-grained features, our results suggest this is not always beneficial, particularly in larger environments. This may be due to the model's increased ability to extract global context information from low resolution input.

TABLE III. ACCURACY OF RESNET-101 MODEL USING DIFFERENT INPUT SIZES ACROSS SIMULATED ENVIRONMENT TEST SETS

| | Cabin | | Warehouse | | Temple | |
|---|---|---|---|---|---|---|
| Input | $P$ (m) | $\theta$(°) | $P$ (m) | $\theta$(°) | $P$ (m) | $\theta$(°) |
| *128×128* | 0.041 | 0.62 | 0.301 | **1.64** | 0.652 | 1.87 |
| *256×256* | 0.034 | 0.52 | 0.346 | 2.19 | **0.56** | **1.57** |
| *512×512* | **0.029** | **0.49** | **0.299** | 2.32 | 0.68 | 2.9 |

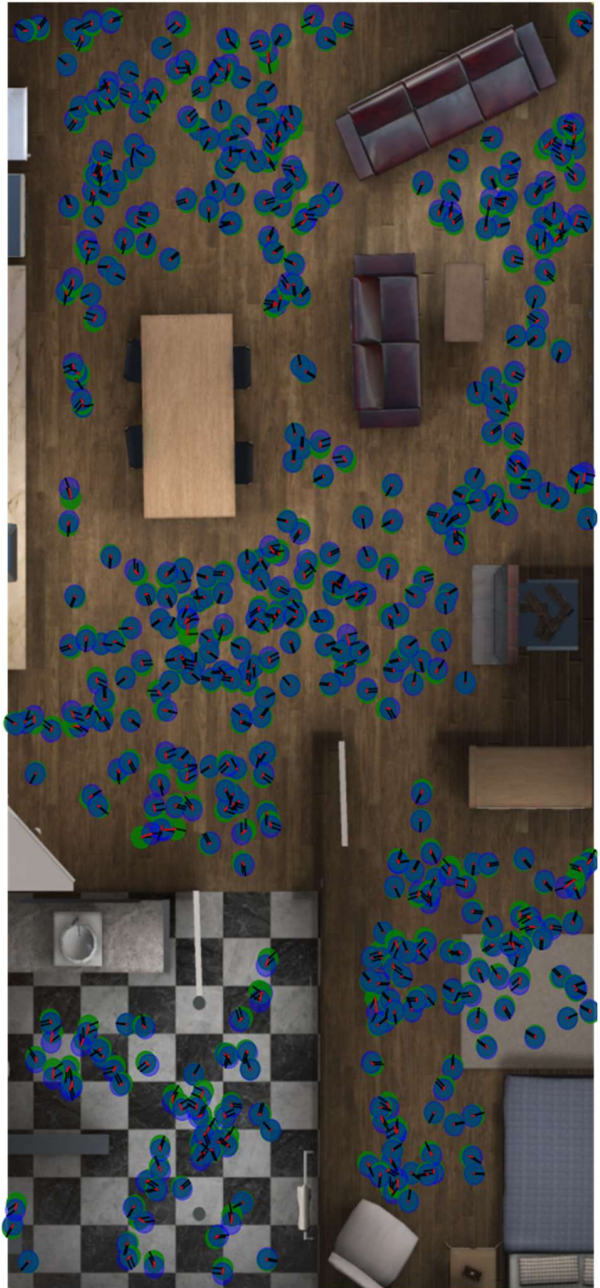

Figure 5. Ground truth (green) and predicted (blue) poses for each sample in the Cabin test set. Black lines show sample orientation, red lines show distance error.

*3) Training Samples*

In Table 4 we evaluate the impact of training set size on model accuracy. In each case we quote the average position and orientation error of a trained Resnet-101 model using input dimensions of 256×256 and a batch size of 32. As would be expected, decreasing the number of training samples has a negative effect on performance, especially in larger environments, however only the mid-sized Warehouse environment benefits from the increase from 100,000 samples to 200,000.

TABLE IV. ACCURACY OF RESNET-101 MODEL USING DIFFERENT AMOUNTS OF TRAINING DATA

|  | Cabin | | Warehouse | | Temple | |
|---|---|---|---|---|---|---|
| **Samples** | *P (m)* | *θ(°)* | *P (m)* | *θ(°)* | *P (m)* | *θ(°)* |
| *200,000* | 0.036 | 0.6 | **0.248** | **1.231** | 0.585 | 1.75 |
| *100,000* | **0.034** | **0.52** | 0.346 | 2.19 | **0.56** | **1.57** |
| *50,000* | 0.04 | 0.66 | 0.45 | 2.043 | 0.73 | 2.61 |
| *20,000* | 0.056 | 1 | 0.716 | 3.156 | 1.062 | 4.15 |

*B. Real Environment*

We evaluate the models trained on our real-world data on a test set of 523 images captured in the same environment, with the error in estimated position and orientation across this test set detailed in Table 5a. In each case we use input dimensions of 256×256, a training set of 32,638 samples and a batch size of 32. Our results demonstrate comparable performance between Resnet variants and VGG19, while MobileNet-v2 generally performs worse. Resnet-50 demonstrates the most accurate orientation prediction, with similar position accuracy to its larger counterparts, leading us to select this model for deployment on our UGV platform.

TABLE V. RESULTS FROM MODELS TRAINED FOR OUR REAL ENVIRONMENT

| Apartment | Position Error (m) | | | | Orientation Error (°) | | | |
|---|---|---|---|---|---|---|---|---|
| Model | μ | σ | Max | RMSE | μ | σ | Max | RMSE |
| *Resnet-18* | 0.093 | 0.070 | **0.671** | 0.117 | 4.853 | 12.78 | 172.3 | 13.67 |
| *Resnet-34* | 0.092 | 0.083 | 1.422 | 0.124 | 3.747 | 7.077 | 95.76 | 8.008 |
| *Resnet-50* | 0.090 | 0.078 | 0.952 | 0.119 | **3.334** | **5.076** | **74.95** | **6.073** |
| *Resnet-101* | 0.092 | **0.069** | 0.685 | **0.115** | 3.634 | 7.026 | 111.6 | 7.910 |
| *Resnet-152* | **0.089** | 0.091 | 1.686 | 0.128 | 3.765 | 9.784 | 178.7 | 10.48 |
| *VGG19* | 0.096 | 0.080 | 0.950 | 0.125 | 3.528 | 5.185 | 80.55 | 6.272 |
| *MobileNet-v2* | 0.112 | 0.117 | 2.050 | 0.162 | 4.320 | 9.053 | 129.9 | 10.03 |

a

| Input | *P (m)* | *θ(°)* |
|---|---|---|
| *128×128* | **0.09** | **3.29** |
| *256×256* | 0.092 | 3.63 |
| *512×512* | 0.103 | 4.29 |

b

| Samples | *P (m)* | *θ(°)* |
|---|---|---|
| *32,638* | **0.092** | 3.63 |
| *6700* | 0.177 | **3.34** |

c

Our simulation results demonstrate that input image size can affect performance significantly, so we evaluate Resnet-101 models trained for different input dimensions, the results of which are listed in Table 5b. Our results show a clear improvement when smaller input images are used, which differs from what was observed with simulation data, suggesting that global context information extracted at lower resolutions may be more important than high-resolution details. We evaluate the effect of training set size by training a Resnet-101 model on a subset of our training data comprising 6700 samples. The results, shown in Table 5c, demonstrate a significant decrease in position accuracy but not in orientation.

Figure 6. visualises the predictions of the trained Resnet-50 model, showing that the majority of predictions are accurate,

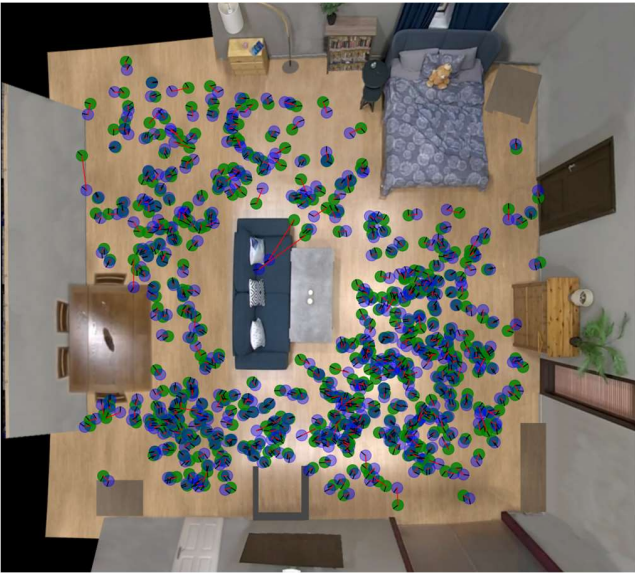

Figure 6. Ground truth (green) and predicted (blue) poses for each sample in the Apartment test set. Black lines show sample orientation, red lines show distance error.

with a small number of outliers. Figure 7 shows the samples with the largest position and orientation errors, of 1m and 75° respectively. While the former image was captured very close to an obstacle and is severely impacted by motion blur and rolling shutter effect, it is not clear what caused the failure in the latter instance.

## C. Model Size and Speed Comparison

For an autonomous mobile platform to effectively navigate in the real world, the model used to estimate position and orientation needs to be capable of running onboard with as low a latency as possible. We evaluate the time taken for each of our models to run inference on a single frame across different input dimensions in three hardware scenarios: Nvidia RTX 3090 [37] workstation GPU, as used for training our models; Nvidia Jetson Xavier [38], a compact embedded GPU platform widely used onboard autonomous systems; and the onboard computer of a UGV, which uses an Intel Core i5 CPU [39] and has no dedicated GPU. Table 6 lists the mean frames per second in each scenario, averaged across 100 input frames, as well as the size of each model.

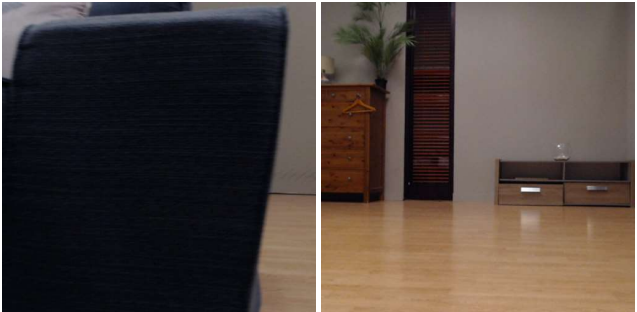

Figure 7. Samples from the Apartment test set that resulted in the largest position (left) and orientation (right) errors.

TABLE VI. SIZE AND SPEED COMPARISON OF THE CNN ARCHITECTURES EVALUATED

|  | Resnet-18 | Resnet-34 | Resnet-50 | Resnet-101 | Resnet-152 | Mobilenet-V2 | VGG19 |
|---|---|---|---|---|---|---|---|
| *Model Size* | 44MB | 83MB | 92MB | 167MB | 228MB | 9MB | 545MB |
| *RTX 3090 Workstation fps* | | | | | | | |
| *512×512* | 484.2 | 272.1 | 211.1 | 110.0 | 74.3 | 288.8 | 92.0 |
| *256×256* | 505.1 | 280.8 | 219.9 | 112.0 | 75.6 | 299.2 | 284.0 |
| *128×128* | 504.9 | 285.3 | 221.1 | 114.5 | 76.1 | 303.8 | 546.5 |
| *Jetson Xavier Embedded GPU fps* | | | | | | | |
| *512×512* | 37.6 | 21.8 | 11.9 | 6.75 | 4.67 | 42.2 | 4.8 |
| *256×256* | 101.3 | 57.6 | 35.0 | 20.6 | 14.5 | 83.1 | 16.2 |
| *128×128* | 154.4 | 87.1 | 63.5 | 35.5 | 23.8 | 81.4 | 18.8 |
| *UGV onboard CPU fps* | | | | | | | |
| *512×512* | 4.4 | 2.6 | 1.5 | 0.89 | 0.62 | 4.9 | 0.5 |
| *256×256* | 15.3 | 8.6 | 5.6 | 3.4 | 2.4 | 22.8 | 1.9 |
| *128×128* | 33.4 | 19.1 | 14.3 | 8.1 | 5.8 | 65.2 | 5.6 |

## D. Live Testing

To demonstrate the viability of neural mapping in real-world robotics scenarios, we deploy a trained model to the onboard computer of a UGV and perform a waypoint navigation task. We first define a list of waypoints that the UGV will aim to traverse, then using the trained model to estimate its pose at each step, the UGV navigates by the procedure laid out in Algorithm 1: For each waypoint $W_{x,y}$, the UGV's pose $P_{x,y,\theta}$ is estimated; If the distance $D$ between $W_{x,y}$, and $P_{x,y}$ is less than predetermined threshold $T_d$, it is determined to have reached the waypoint; If $D > T_d$, the heading $H$ from $W_{x,y}$ to $P_{x,y}$ is computed. If the difference between $H$ and $P_\theta$ is greater than predetermined threshold, $T_a$, rotation commands are sent to the robot until the odometry data obtained from the onboard IMU and wheel encoders suggest that it has rotated to within $T_a$ of $H$; $P_{x,y,\theta}$ is estimated again, and if necessary another rotation step is performed, otherwise commands to move forwards are sent to the UGV until the odometry shows it has moved either 1m or $D$, whichever is less, at which point $P_{x,y,\theta}$ is estimated again and the process repeats.

We set a loop of 8 waypoints around the apartment environment, and set $T_a$ to 5° and $T_d$ to 0.5m, as we find that smaller thresholds cause the UGV (with dimensions of approximately 1m × 0.7m) to spend a lot of time making minor adjustments. The UGV's ultrasonic positioning system is used to obtain ground truth pose for evaluation, however it is not used for navigation.

Using a Resnet-50 model with input size of 256×256, the UGV successfully navigated through all 8 waypoints, with the route visualized in Figure 8 (the route starts where the green and blue lines begin on the left side of the image). We can see that position error increased in the top section of the map, however when the UGV reached the left section it was able to predict accurately again, illustrating an advantage over relative pose approaches, such as SLAM or visual odometry -based techniques, wherein errors can compound over subsequent frames. The mean closest distance the UGV reached from each waypoint is 31.1cm, although the mean distance based on estimated pose was 18.3cm, meaning the model predicted it was closer than it actually was.

**Algorithm 1** Pseudocode for waypoint navigation
1: **for** $W_{x,y}$ in waypoint list **do**
2:    $P_{x,y,\theta} \leftarrow$ estimated current pose
3:    $D \leftarrow$ distance between $W_{x,y}$ and $P_{x,y}$
4:    **while** $D >$ threshold $T_d$ **do**
5:       $H \leftarrow$ heading from $P_{x,y}$ to $W_{x,y}$
6:       **while** $|P_\theta - H| >$ threshold $T_a$ **do**
7:          $R \leftarrow H - P_\theta$
8:          $A \leftarrow 0$
9:          **while** $|R - A| >$ threshold $T_a$ **do**
10:             send rotation command to robot
11:             $A \leftarrow$ angular odometry from robot
12:          $P_{x,y,\theta} \leftarrow$ estimate current pose
13:       $M \leftarrow \min(D, 1)$
14:       $L \leftarrow 0$
15:       **while** $|M - L| >$ threshold $T_d$ **do**
16:          send movement command to robot
17:          $L \leftarrow$ linear odometry from robot
18:       $P_{x,y,\theta} \leftarrow$ estimate current pose
19:       $D \leftarrow$ distance between $W_{x,y}$ and $P_{x,y}$

## V. CONCLUSION

In this work we have demonstrated the capability for an efficient neural mapping pose estimation approach to be effectively deployed onboard an unmanned ground vehicle. We have created a novel dataset encompassing simulated and real environments, and demonstrated that by reducing the complexity of the pose estimation problem to 2 dimensions and increasing the quantity of training samples, accurate estimation of position and orientation is possible in real time.

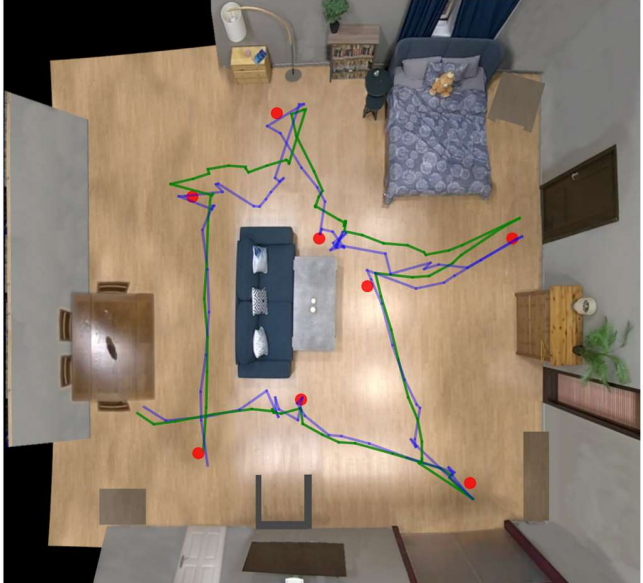

Figure 8. The route taken by the UGV during our live test. Waypoints are marked in red, the route according to the poses predicted by the model is marked in blue, while the ground truth route from the ultrasonic positioning system is marked in green.